\def\year{2020}\relax
\documentclass[letterpaper]{article} 
\usepackage{aaai20}  
\usepackage{times}  
\usepackage{helvet} 
\usepackage{courier}  
\usepackage[hyphens]{url}  
\usepackage{graphicx} 
\usepackage{graphicx}  
\newcommand{\etal}{\textit{et al}.~}
\newcommand{\ie}{\textit{i}.\textit{e}.~}
\newcommand{\ieno}{\textit{i}.\textit{e}.}
\newcommand{\eg}{\textit{e}.\textit{g}.~}
\newcommand{\egno}{\textit{e}.\textit{g}.} 
\newcommand{\etc}{\textit{etc}.}
\newcommand{\etcno}{\textit{etc}} 
\newcommand{\triIDC}{\text{Triplet ReID constraints }}

\usepackage{algorithm}
\usepackage{algorithmicx}
\usepackage{algpseudocode}
\usepackage{amsmath}
\usepackage{booktabs}       
\usepackage{multirow}
\usepackage{hyperref}

\usepackage{enumitem}
\usepackage{amsfonts}

\title{Semantics-Aligned Representation Learning for Person Re-identification}

\author{
Xin Jin$^{1}$\thanks{This work was done when Xin Jin was an intern at MSRA.} \qquad Cuiling Lan$^{2}$\thanks{Corresponding Author.} \qquad Wenjun Zeng$^{2}$ \qquad Guoqiang Wei$^{1}$ \qquad Zhibo Chen$^{1\dag}$ \\
University of Science and Technology of China$^{1}$ \qquad Microsoft Research Asia$^{2}$ \\
{\tt\small \{jinxustc,wgq7441\}@mail.ustc.edu.cn\quad \{culan, wezeng\}@microsoft.com\quad chenzhibo@ustc.edu.cn}}

\begin{document}

\maketitle

\begin{abstract}
Person re-identification (reID) aims to match person images to retrieve the ones with the same identity. This is a challenging task, as the images to be matched are generally semantically misaligned due to the diversity of human poses and capture viewpoints, incompleteness of the visible bodies (due to occlusion), \etcno. In this paper, we propose a framework that drives the reID network to learn semantics-aligned feature representation through delicate supervision designs. Specifically, we build a Semantics Aligning Network (SAN) which consists of a base network as encoder (SA-Enc) for re-ID, and a decoder (SA-Dec) for \emph{reconstructing/regressing the densely semantics aligned full texture image}. We jointly train the SAN under the supervisions of person re-identification and aligned texture generation. Moreover, at the decoder, besides the reconstruction loss, we add Triplet ReID constraints over the feature maps as the perceptual losses. The decoder is discarded in the inference and thus our scheme is computationally efficient. Ablation studies demonstrate the effectiveness of our design. We achieve the state-of-the-art performances on the benchmark datasets CUHK03, Market1501, MSMT17, and the partial person reID dataset Partial REID. Code for our proposed method is available at: \url{https://github.com/microsoft/Semantics-Aligned-Representation-Learning-for-Person-Re-identification}.
\end{abstract}

\section{Introduction}

Person re-identification (reID) aims to identify/match persons in different places, times, or camera views. There are large variations in terms of the human poses, capturing view points, incompleteness of the bodies (due to occlusion). These result in \emph{semantics misalignment} across 2D images which makes reID challenging \cite{shen2015person,varior2016siamese,subramaniam2016deep,su2017pose,zheng2017pose,zhang2017alignedreid,yao2017deep,li2017learning,zhao2017spindle,wei2017glad,zheng2018pedestrian,ge2018fd,suh2018part,qian2018pose,zhang2019DSA}.

\begin{figure}[t]
  \centerline{\includegraphics[width=1.0\linewidth]{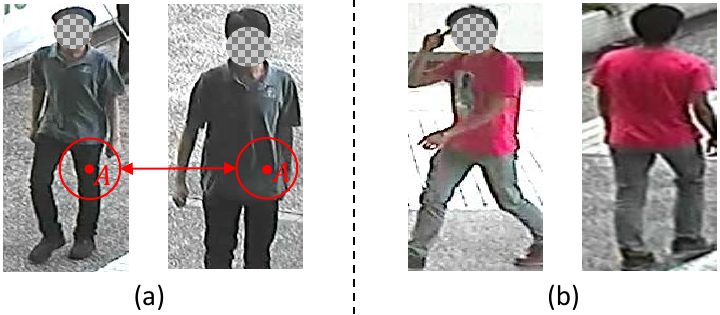}}
  \caption{Challenges in person reID: (a) Spatial misalignment; (b) Inconsistency of the visible body regions/semantics.}
  \centering
\label{fig:motivation}
\end{figure}

\emph{Semantics misalignment} can be interpreted from two aspects. (1) Spatial semantics misalignment: the same spatial position across images may correspond to different semantics of human body or even different objects. As the example in Figure \ref{fig:motivation} (a) shows, the spatial position $A$ which corresponds to person leg in the first image corresponds to person abdomen in the second image. (2) Inconsistency of visible body regions/semantics: since a person is captured through a 2D projection, only a portion of the 3D surface of a person is visible/projected in an image. The visible body regions/semantics across images are not consistent. As shown in Figure \ref{fig:motivation}(b), front side of a person is visible in one image and invisible in another one.

\textbf{Alignment:} Deep learning methods can deal with such diversities and misalignment to some extent but it is not enough. In recent years, many approaches explicitly exploit human pose/landmark information to achieve coarse alignment and they have demonstrated their superiority for person reID \cite{su2017pose,zheng2017pose,yao2017deep,li2017learning,zhao2017spindle,wei2017glad,suh2018part}. During the inference, these part detection sub-networks are usually required which increases the computational complexity. Besides, the body-part alignment is coarse and there is still spatial misalignment within the parts \cite{zhang2019DSA}. To achieve fine-granularity spatial alignment, based on estimated dense semantics \cite{guler2018densepose}, Zhang \etal warp the input person image to a canonical UV coordinate system to have densely semantics aligned images as inputs for reID \cite{zhang2019DSA}. However, the invisible body regions result in many holes in the warped images and thus the inconsistency of visible body regions across images. How to better solve the dense semantics misalignment is still an open problem.

\begin{figure}
  \centerline{\includegraphics[width=1.0\linewidth]{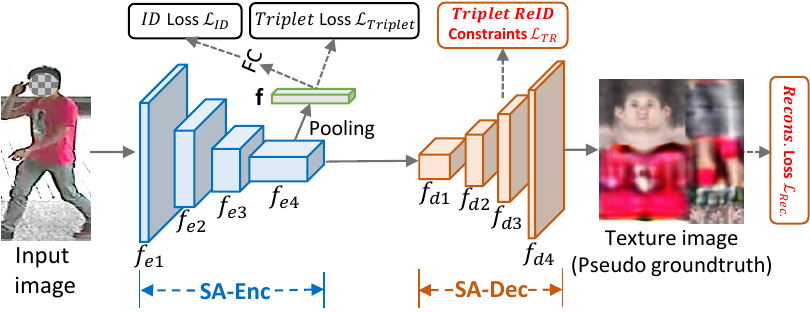}}
  \caption{Illustration of the proposed Semantics Aligning Network (SAN), which consists of a base network as encoder (SA-Enc) and a decoder sub-network (SA-Dec). The reID feature vector $\rm \textbf{f}$ is obtained by average pooling the  feature map $f_{e4}$ of the SA-Enc, followed by the reID losses of $\mathcal{L}_{ID}$ and $\mathcal{L}_{Triplet}$. To encourage the encoder learning semantically aligned features, the SA-Dec is followed which regresses the densely semantically \emph{aligned full} texture image with the pseudo groundtruth supervision $\mathcal{L}_{Rec.}$. The pseudo groundtruth generation is described in Sec.~\ref{3.1} without shown here. At the decoder, \triIDC $\mathcal{L}_{TR}$ are added as the high level perceptual metric. We use ResNet-50 with four residual blocks as our SA-Enc. In inference, the SA-Dec is discarded.}  
  \centering
\label{fig:flowchart}
\end{figure}

\textbf{Our work:} We intend to fully address the semantics misalignment problems in both aspects. We achieve this by proposing a simple yet powerful Semantics Aligning Network (SAN). Figure \ref{fig:flowchart} shows the overall framework of the SAN, which introduces an aligned texture generation sub-task, with densely semantics aligned texture image (see examples in Figure \ref{fig:texture}) as supervision. Specifically, SAN consists of a base network as encoder (SA-Enc), and a decoder sub-network (SA-Dec). The SA-Enc can be any baseline network used for person reID (\eg ResNet-50 \cite{he2016deep}), which outputs a feature map $f_{e4}$ of size $h\times w \times c$. The reID feature vector $\rm \textbf{f} \in \mathbb{R}^c$ is then obtained by average pooling the feature map $f_{e4}$, followed by the reID losses. To encourage the SA-Enc to learn semantically aligned features, the SA-Dec is introduced and used to regress/generate the densely semantically aligned full texture image (also referred to as texture image for short) with pseudo groundtruth supervision. We exploit a synthesized dataset for learning pseudo groundtruth texture image generation. This framework enjoys the benefit of dense semantics alignment but without increasing the complexity of inference since the decoder SA-Dec is discarded in inference.

Our main contributions are summarized as follows.
\begin{itemize}
\item We propose a simple yet powerful framework for solving the misalignment challenge in person reID without increasing computational cost in inference. 
\item A semantics alignment constraint is delicately introduced by empowering the encoded feature map with \emph{aligned full} texture generation capability.
\item At the SA-Dec, besides the reconstruction loss, we propose \triIDC over the feature maps as the perceptual metric.
\item There is no groundtruth aligned texture image for the person reID datasets. We address this by generating pseudo groundtruth texture images by leveraging synthesized data with person image and aligned texture image pairs (see Figure \ref{fig:texture}).
\end{itemize}

Our method achieves the state-of-the-art performance on the benchmark datasets CUHK03 \cite{li2014deepreid}, Market-1501 \cite{zheng2015scalable}, MSMT17 \cite{wei2018person}, Partial REID \cite{zheng2015partial}.

\section{Related Work}\label{sec2}

Person reID based on deep neural networks has made great progress in recent years. Due to the variations in poses, viewpoints, incompleteness of the visible bodies (due to occlusion), \etc, across the images, semantics misalignment is still one of the key challenges.

\textbf{Alignment with Pose/Part Cues for ReID:} To address the spatial semantics misalignment, most of the previous approaches make use of external cues such as pose/part \cite{li2017learning,yao2017deep,zhao2017spindle,kalayeh2018human,zheng2017pose,su2017pose,suh2018part}. Human landmark (pose) information can help align body regions across images. Zhao \etal \cite{zhao2017spindle} propose a human body region guided Splindle Net, where a body region proposal sub-network (trained with the human pose dataset) is used to extract the body regions, \egno, head-shoulder, arm region. The semantic features from different body regions are separately captured thus the body part features can be aligned across images. Kalayeh \etal \cite{kalayeh2018human} integrate a human semantic parsing branch in their network for generating probability maps associated to different semantic regions of human body, \egno, head, upper-body. Based on the probability maps, the features from different semantic regions of human body are aggregated separately to have part aligned features. Qian \etal \cite{qian2018pose} propose to make use of GAN model to synthesize realistic person images of eight canonical poses for matching. However, these approaches usually require pose/part detection or image generation sub-networks, and extra computational cost in inference. Moreover, the alignment based on pose is coarse without considering the finer grained alignment within a part across images.

{Zhang \etal \cite{zhang2019DSA} exploit the dense semantics from DensePose \cite{alp2018densepose} rather than the coarse pose for reID. Their network consists of two streams in training: a main stream takes the original image as input while the other stream learns features from the warped images for regularizing the feature learning of the main stream. However, the invisible body regions result in many holes in the warped images and  inconsistency of visible body regions across images, which could hurt the learning efficiency. Moreover, there is a lack of more direct constraints to enforce the alignment. The design of efficient frameworks for dense semantics alignment is still under-explored. In this paper, we propose an elegant framework which adds direct constraints to encourage dense semantics alignment in feature learning.}

\begin{figure}
  \centerline{\includegraphics[width=1.0\linewidth]{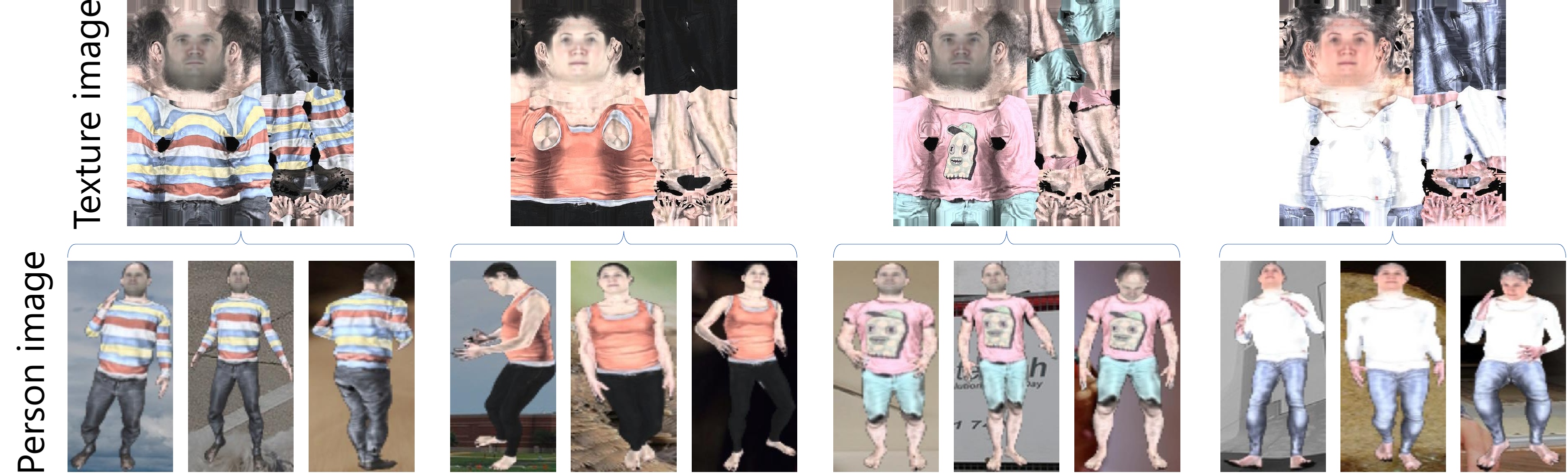}}
  \caption{Examples of texture images (first row) and the corresponding synthesized person images with different poses, viewpoints, and backgrounds (second row). A texture image represents the full texture of the 3D human surface in a surface-based canonical coordinate system (UV space). Each position (u,v) corresponds to a unique semantic identity. For person images of different persons/poses/viewpoints (in the second row), their corresponding texture images are densely semantically aligned.}
  \centering
\label{fig:texture}
\end{figure}

\begin{figure}
  \centerline{\includegraphics[width=1.0\linewidth]{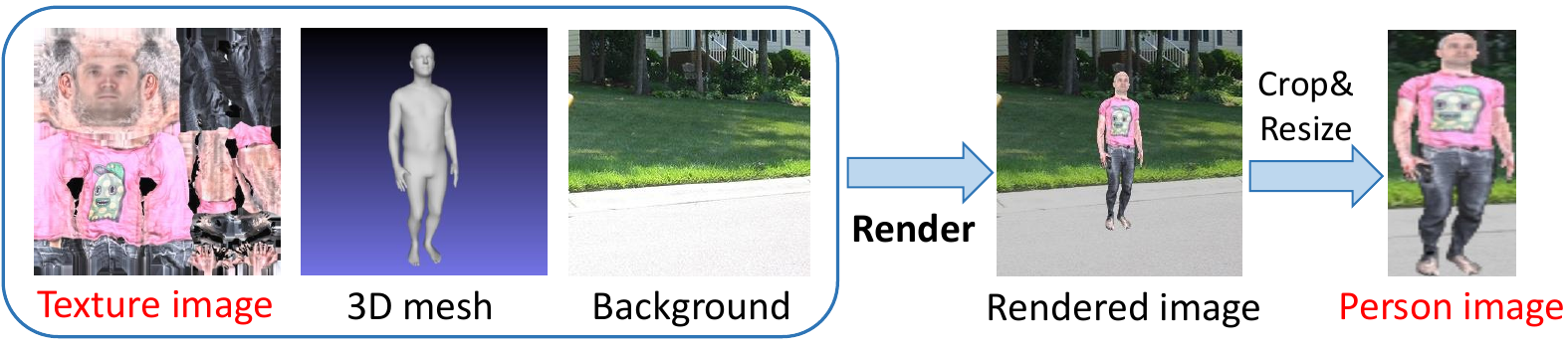}} 
  \caption{Illustration of the generation of synthesized person image to form a (\emph{person image}, \emph{texture image}) pair. Given a texture image, a 3D mesh, a background image, and rendering parameters, we can obtain a 2D person image through the rendering.} 
  \centering
\label{fig:syn}
\end{figure}

\textbf{Semantics Aligned Human Texture:} A human body could be represented by a 3D mesh (\eg Skinned Multi-Person Linear Model, SMPL \cite{loper2015smpl}) and a texture image \cite{varol2017learning,hormann2007mesh} as illustrated in Figure \ref{fig:syn}. Each position on the 3D body surface has a semantic identity (identified by a 2D coordinate (u,v) in the canonical UV space) and a texture representation (\eg RGB pixel value) \cite{guler2018densepose,guler2017densereg}. A texture image on the UV coordinate system (\ieno, surface-based coordinate system) represents the \emph{aligned full} texture of the 3D surface of the person. Note that the texture images across different persons are densely semantically aligned (see Figure \ref{fig:texture}). In \cite{guler2018densepose}, a dataset with labeled dense semantics (\ie DensePose) is established and a CNN-based system is designed to estimate DensePose from person images. Neverova \etal \cite{neverova2018dense} and Wang \etal \cite{wang2019re} leverage the aligned texture image to synthesize person image of another pose or view. Yao \etal \cite{yao2019densebody} propose to regress the 3D human body ((x,y,z) coordinates in 3D space) in the semantics aligned UV space, with the RGB person image as the input to the CNN.

Different from all these works, we leverage the densely semantically aligned full texture image to address the misalignment problem in person reID. We use them as direct supervisions to drive the reID network to learn semantics aligned features.

\section{The Semantics Aligning Network (SAN)}\label{sec3}

To address the cross image misalignment challenge caused by human pose, capturing viewpoint variations, and the incompleteness of the body surface (due to the occlusion when projecting 3D person to 2D person image), we propose a Semantics Aligning Network (SAN) for robust person reID, in which densely semantically aligned full texture images are taken as supervision to drive the learning of semantics aligned features.

The proposed framework is shown in Figure \ref{fig:flowchart}. It consists of a base network as encoder (SA-Enc) for reID, and a decoder sub-network (SA-Dec) (see Sec. \ref{3.2}) for generating densely semantically aligned full texture image with supervision. This encourages the reID network to learn semantics aligned feature representation. Since there is no groundtruth texture image of 3D human surface for the reID datasets, we use our synthesized data based on \cite{varol2017learning} to train SAN (with reID supervisions removed) which is then used to generate pseudo groundtruth texture images for the reID datasets (see Sec. \ref{3.1}).

The reID feature vector $\rm \textbf{f}$ is obtained by average pooling the last layer feature map $f_{e4}$ of the SA-Enc, followed by the reID losses. The SA-Dec is added after the last layer of the SA-Enc to regress densely semantically aligned texture image, with the (pseudo) groundtruth texture supervision. In the SA-Dec, \triIDC    are further incorporated at different layers/blocks as the high level perceptual metric to encourage identity preserving reconstruction. During inference, the SA-Dec is discarded.

\subsection{Densely Semantically Aligned Texture Image}\label{3.1}

\textbf{Background:} The person texture image in the surface-based coordinate system (UV space) is widely used in the graphics field \cite{hormann2007mesh}. Texture images for different persons/viewpoints/poses are densely semantically aligned, as illustrated in Figure \ref{fig:texture}. Each position (u,v) corresponds to a unique semantic identity on the texture image, \egno, the pixel on the right bottom of the texture image corresponds to some semantics of a hand. Besides, a texture image contains all the texture of the full 3D surface of a person. In contrast, only a part of the surface texture is visible/projected on a 2D person image.   

\textbf{Motivation:} We intend to leverage such aligned texture images to drive the reID network to learn semantics aligned features. For different input person images, the corresponding texture images are well semantics aligned. First, for the same spatial positions on different texture images, the semantics are the same. Second, for person images with different visible semantics/regions, their texture images are semantics consistent/aligned since each one contains the full texture/information of the 3D person surface.

\textbf{Pseudo groundtruth Texture Images Generation:} For the images in the reID datasets, however, there are no groundtruth aligned full texture images. We propose to train the SAN using our synthesized data to enable the generation of a pseudo groundtruth texture image for each image in the reID datasets. We can leverage a CNN-based network to generate pseudo groundtruth texture images. In this work, we reuse the proposed SAN (with the reID supervisions removed) as the network (see Figure \ref{fig:flowchart}), which we refer to as SAN-PG (Semantics Aligning Network for Pseudo Groundtruth Generation) for differentiation. Given an input person image, the SAN-PG outputs predicted texture image as the pseudo groundtruth. 

To train the SAN-PG, we synthesize a Paired-Image-Texture dataset (PIT dataset), based on SURREAL dataset \cite{varol2017learning}, for the purpose of providing the image pairs, \ieno, the \emph{person image} and \emph{its texture image}. The texture image stores the RGB texture of the \emph{full} person 3D surface. As illustrated in Figure \ref{fig:syn}, given a texture image, a 3D mesh/shape, and a background image, a 2D projection of a 3D person can be obtained by rendering \cite{varol2017learning}. We can control the pose and body form of the person, and projection viewpoint, through changing the parameters of 3D mesh/shape model (\ie SMPL \cite{loper2015smpl}) and the rendering parameters. Note that we do not include identity information in the PIT dataset.

To generate the PIT dataset with paired person images and texture images, in particular, we use 929 (451 for female and 478 for male) raster-scanned texture maps provided by the SURREAL dataset \cite{varol2017learning} to generate the \emph{person image} and \emph{texture image} pairs. These texture images are aligned with the SMPL default two-dimensional UV coordinate space (UV space). The same uv coordinate value corresponds to the same semantics. We generate 9,290 different meshes of diverse poses/shapes/viewpoints, by using SMPL body model \cite{loper2015smpl} parameters inferred by HMR \cite{kanazawa2018end} from the person images of the COCO dataset \cite{lin2014microsoft}. For each texture map, we assign 10 different meshes and render these 3D meshes with the texture image by Neural Render \cite{kato2018neural}. Then we obtain in total 9,290 different synthesized (\emph{person image}, \emph{texture image}) pairs. To simulate real-world scenes, the background images for rendering are randomly sampled from COCO dataset \cite{lin2014microsoft}. Each synthetic person image is centered on a person with resolution 256$\times$128. The resolution of the texture images is 256$\times$256.

\textbf{Discussion:} The texture images which we use for supervisions have three major advantages. 1) They are spatially aligned in terms of the dense semantics of a person surface and thus can guide the reID network to learn semantics aligned representation. 2) A texture image containing the \emph{full} 3D surface of a person can guide the reID network to learn more comprehensive representation of a person. 3) They represent the textures of the human body surface and thus naturally eliminate the interference of diverse background scenes.

There are also some limitations of the current pseudo groundtruth texture image generation process. 1) There is a domain gap between synthetic 2D images (in the PIT dataset) and real-world captured images where the synthetic person is not very realistic. 2) The number of texture images provided by SURREAL \cite{varol2017learning} is not large (\ie 929 in total) which may constraint the diversity of the data in our synthesized dataset. 3) On SURREAL, all faces in the texture image are replaced by an average face of either man or woman \cite{lin2014microsoft}. We leave it as future work to address these limitations. Even with such limitations, our scheme achieves significant performance improvement over the baseline on person reID.

\subsection{SAN and Optimization}\label{3.2} 

As illustrated in Figure \ref{fig:flowchart}, the SAN consists of an encoder SA-Enc for person reID, and a decoder SA-Dec which enforces constraints over the encoder by requiring the encoded features to be able to predict/regress the semantically aligned full texture images.

\textbf{SA-Enc:} We can use any baseline network used in person reID (\eg ResNet-50 \cite{sun2017beyond,zhang2017alignedreid,zhang2019DSA}) as the SA-Enc. In this work, we similarly use ResNet-50 and it consists of four residual blocks. The output feature map of the fourth block $f_{e4} \in \mathbb{R}^{ h \times w \times c}$ is spatially average pooled to get the feature vector (${\rm \textbf{f}} \in \mathbb{R}^c$), which is the reID feature for matching. 

For the purpose of reID, on the feature vector ${\rm \textbf{f}}$, we add the widely-used identification loss ($ID$ Loss) $\mathcal{L}_{ID}$, \ieno, the cross entropy loss for identification classification, and the ranking loss of triplet loss with batch hard mining \cite{hermans2017defense} ($Triplet$ Loss) $\mathcal{L}_{Triplet}$ as the loss functions in training. 

\textbf{SA-Dec:} To encourage the encoder features to learn semantics aligned features, we add a decoder SA-Dec after the fourth block ($f_{e4}$) of the encoder to regress the densely semantically aligned texture images, supervised by the (pseudo) groundtruth texture images. A reconstruction loss $\mathcal{L}_{Rec.}$ is introduced to minimize $L$1 differences between the generated texture image and its corresponding (pseudo) groundtruth texture image. 





\textbf{{Triplet ReID constraints at SA-Dec:}} Besides the capability of reconstructing the texture images optimized/measured by the $L$1 distance, we also expect the features in the decoder inherit the capability of distinguishing different identities. Wang \etal \cite{wang2019re} use reID network as the perceptual supervision to generate person image, which judges whether the generated person image and the real image have the same identity. Different from \cite{wang2019re}, in considering that the features at each layer of the decoder are spatially semantically aligned across images, we measure the feature distance for each spatial position rather than on the final globally pooled feature. We introduce $Triplet$ $ReID$ constraints to minimize the $L$2 differences between the features of the same identity and maximize those of different identities. Specially, for a sample $a$ in a batch, we can randomly select a positive sample $p$ (with the same identity) and a negative sample $n$. The Triplet ReID constraint/loss over the output feature map of the $l^{th}$ block of the SA-Dec is defined as  
\begin{equation}
\begin{aligned}
\mathcal{L}_{TR}^l = max(\frac{1}{h_l\times w_l}||f_{dl}(x_{l}^{a})-f_{dl}(x_{l}^{p})||_{2}^{2} - \\
\frac{1}{h_l\times w_l}||f_{dl}(x_{l}^{a}-f_{dl}(x_{l}^{n})||_{2}^{2}+ m, 0),
\label{eq:2}
\end{aligned}
\end{equation}
where $h_l \times w_l$ is the resolution of feature map with $c_l$ channels, $f_{dl}(x_{l}^{a}) \in \mathbb{R}^{h_l \times w_l \times c_l}$ denotes the feature map of sample $a$.  $||f_{dl}(x_{l}^{a})-f_{dl}(x_{l}^{p})||_{2}^{2} = \sum_{i=1}^{h_l} \sum_{j=1}^{w_l} || f_{dl}(x_{l}^{a})(i,j,:) - f_{dl}(x_{l}^{p})(i,j,:) ||_{2}^{2}$ with $f_{dl}(x_{l}^{a})(i,j,:)$ denotes the feature vector of $c_l$ channels at spatial position $(i,j)$. The margin parameter $m$ is set to 0.3 experimentally

\textbf{Training Scheme:} There are two steps for training our proposed SAN framework for reID:

\emph{Step-1}, we train a network for the purpose of generating pseudo groundtruth texture images for any given input person image. For simplicity, we reuse a simplified SAN (\ieno, SAN-PG) which consists of the SA-Enc and SA-Dec, but with only the reconstruction loss $\mathcal{L}_{Rec.}$. We train the SAN-PG with our synthesized PIT dataset. The SAN-PG model is then used to generate pseudo groundtruth texture image for reID datasets (such as CUHK03 \cite{li2014deepreid}). 

\emph{Step-2}, we train the SAN for both reID and aligned texture generation. The pre-trained weights of the SAN-PG are used to initialize the SAN. One alternative is to use only the reID dataset for training SAN, where the pseudo groundtruth texture images are used for supervision and all the losses are added. The other strategy is to iteratively use the reID dataset and the synthesized PIT dataset during training. We find the later solution gives superior results because the groundtruth texture images for the synthesized PIT dataset have higher quality than that of reID dataset. The overall loss  $\mathcal{L}$ consists of the $ID$ Loss $\mathcal{L}_{ID}$, the $Triplet$ Loss $\mathcal{L}_{Triplet}$, the reconstruction loss $\mathcal{L}_{Rec.}$, and the $Triplet$ $ReID$ constraint $\mathcal{L}_{TR}$, \ieno, $\mathcal{L}$ =  $\lambda_1\mathcal{L}_{ID}$ +  $\lambda_2\mathcal{L}_{Triplet}$ +  $\lambda_3\mathcal{L}_{Rec.}$ +  $\lambda_4\mathcal{L}_{TR}$. For a batch of reID data, we experimentally set $\lambda_1$ to $\lambda_4$ as 0.5, 1.5, 1, 1. For a batch of synthesized data, $\lambda_1$ to $\lambda_4$ are set to 0, 0, 1, 0 where the reID losses and \triIDC (losses) are not used.

\section{Experiment}\label{experiment}

\subsection{Datasets and Evaluation Metrics}

We conduct experiments on six benchmark person reID datasets, including CUHK03 \cite{li2014deepreid}, Market1501 \cite{zheng2015scalable}, DukeMTMC-reID \cite{zheng2017unlabeled}, the large-scale MSMT17 \cite{wei2018person}, and two challenging partial person reID datasets of Partial REID \cite{zheng2015partial} and Partial-iLIDS \cite{he2018deep}

We follow the common practices and use the cumulative matching characteristics (CMC) at Rank-k, $k$ = 1, 5, 10, and mean average precision (mAP) to evaluate the performance.

\subsection{Implementation Details}

We use ResNet-50 \cite{he2016deep} (which are widely used in some re-ID systems \cite{sun2017beyond,zhang2019DSA}) to build our SA-Enc. We also take it as our baseline (Baseline) with both ID loss and triplet loss.
Similar to \cite{sun2017beyond,zhang2019DSA}, the last spatial down-sample operation in
the last \emph{Conv} layer is removed. We build a light weight decoder SA-Dec by simply stacking 4 residual up-sampling blocks with about 1/3 parameters of the SA-Enc. This facilitates our model training using only a single GPU.

\subsection{Ablation Study}

We perform comprehensive ablation studies to demonstrate the effectiveness of the designs in the SAN framework, on the datasets of CUHK03 (labeled bounding box setting) and Market-1501 (single query setting).

\textbf{Effectiveness of Dense Semantics Alignment.} In Table \ref{tab:compare-baseline}, \textbf{\emph{SAN-basic}} denotes our basic semantics aligning model which is trained with the supervision of the pseudo groundtruth texture images with loss of $\mathcal{L}_{Rec.}$, the reID losses $\mathcal{L}_{ID}$ and $\mathcal{L}_{Triplet}$. \textbf{\emph{SAN w/$\mathcal{L}_{TR}$}} denotes that the \triIDC at the SA-Dec is added on top of the \emph{SAN-basic}. \textbf{\emph{SAN w/syn. data}} denotes that the (\emph{person image}, \emph{texture image}) pairs of our PIT dataset is also used in training the SAN on top of the \emph{SAN-basic} network. \textbf{\emph{SAN}} denotes our final scheme with both the \triIDC and the groundtrth texture image supervision from the PIT on top of the \emph{SAN-basic} network. 

\begin{figure*}
  \centerline{\includegraphics[width=1.0\linewidth]{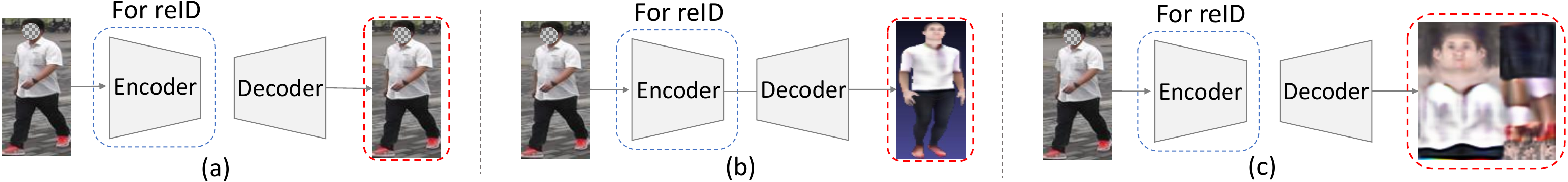}}
  \caption{The same encoder-decoder networks but with different reconstruction objectives of reconstructing the (a) input image, (b) pose aligned person image, and (c) texture image, respectively.}
  \centering
\label{fig:exp1}
\end{figure*}

\begin{table}[t]
  \centering
  \footnotesize
  \tabcolsep=7pt
  \caption{Comparisons (\%) of our SAN and baseline.} 
    \begin{tabular}{ccccc}
    \toprule
    \multirow{2}[4]{*}{Model} & \multicolumn{2}{c}{CUHK03(L)} & \multicolumn{2}{c}{Market1501} \\
\cmidrule{2-5}          & Rank-1 & mAP   & Rank-1 & mAP \\
    \midrule
    Baseline (ResNet-50) & 73.7  & 69.8 & 94.1  & 83.2 \\
    \hline
    SAN-basic           & 77.9  & 73.7 & 95.1  & 85.8 \\
    SAN w/ $\mathcal{L}_{TR}$ & 78.9  & 74.9 & 95.4  & 86.9  \\
    SAN w/ syn. data & 78.8 & 75.8 & 95.7 & 86.8 \\
    SAN & \textbf{80.1}  & \textbf{76.4} & \textbf{96.1}  & \textbf{88.0}  \\
    \bottomrule
    \end{tabular}%
  \label{tab:compare-baseline}%
\end{table}%

We have the following observations/conclusions. \textbf{1)} Thanks to the drive to learn semantics aligned features, our \emph{SAN-basic} significantly outperforms the baseline scheme by about 4\% in both Rank-1 and mAP accuracy on CUHK03. \textbf{2)} The introduction of high-level \triIDC ($\mathcal{L}_{TR}$) as the perceptual loss can regularize the feature learning and it brings about additional 1.0\% and 1.2\% improvements in Rank-1 and mAP accuracy on CUHK03. Note that we add them after each block of the first three blocks in the SA-Dec. \textbf{3)} The use of the synthesized PIT dataset (syn. data) with the input image and groundtruth texture image pairs for training the SAN remedies the imperfection of the generated pesudo groundtruth texture images (with errors/noise/blurring). It improves the performance over \emph{SAN-basic} by 0.9\% and 2.1\% in Rank-1 and mAP accuracy. \textbf{4)} Our final scheme \emph{SAN} significantly outperforms the baseline, \ieno, by \textbf{6.4\%} and \textbf{6.6\%} in Rank-1 and mAP accuracy on CUHK03, but with the same inference complexity.
On Market1501, even though the baseline performance is already very high, our \emph{SAN} achieves 2.0\% and 4.8\% improvement in Rank-1 and mAP.

\textbf{Different Reconstruction Guidance.} We study the effect of using different reconstruction guidance and show results in Table \ref{tab:supervision}. We design another two schemes for comparisons. For the same input image, the three schemes use the same encoder-decoder networks (the same network as SAN-basic) but to reconstruct (a) the input person image, (b) pose aligned person image, and (c) proposed texture image (see Figure \ref{fig:exp1}). To have pose aligned person image as supervision, during synthesizing the PIT dataset, for each projected person image, we also synthesized a person image of a given fixed pose (frontal pose here). Thus, the pose aligned person images are also semantically aligned. In this case, only partial texture (frontal body regions) of the full 3D surface texture is retained with information loss. {In addition, corresponding to (b), we also use the pose aligned person images generated by PN-GAN \cite{qian2018pose} as the reconstruction guidance and get \emph{Enc-Dec rec.~PN-GAN pose}.}

\begin{table}[t]
  \centering
  \footnotesize
  \tabcolsep=2.6pt
  \caption{Performance (\%) comparisons of the same encoder-decoder networks but with different reconstruction objectives of reconstructing the input image, pose aligned person image, and texture image respectively.} 
    \begin{tabular}{ccccc}
    \toprule
    \multirow{2}[4]{*}{Model} & \multicolumn{2}{c}{CUHK03(L)} & \multicolumn{2}{c}{Market1501} \\
\cmidrule{2-5}          & Rank-1 & mAP   & Rank-1 & mAP \\
    \midrule
    Baseline (ResNet-50) & 73.7  & 69.8 & 94.1  & 83.2 \\
    \hline
    Enc-Dec rec.~input   & 74.4  & 70.8  & 94.3  & 84.0  \\
    Enc-Dec rec.~pose  & 75.8  & 72.0 & 94.4  & 84.5 \\
    Enc-Dec rec.~PN-GAN pose  & 76.1  & 72.6  & 94.3 & 84.7  \\
    Enc-Dec rec.~texture~(SAN-basic)          & \textbf{77.9}  & \textbf{73.7} & \textbf{95.1}  & \textbf{85.8} \\
    \bottomrule
    \end{tabular}%
  \label{tab:supervision}%
\end{table}%

From Table \ref{tab:supervision}, we have the following observations/conclusions. 1) The addition of a reconstruction sub-task helps improve the reID performance which encourages the encoded feature to preserve more original information. \emph{Enc-Dec rec.~input} improves the performance of the baseline by 0.7\% and 1.0\% in Rank-1 and mAP accuracy. However, the input images (and their reconstructions) are not semantically aligned across images. 2) \emph{Enc-Dec rec.~pose} and \emph{Enc-Dec rec.~PN-GAN pose} both enforce the supervision to be \emph{pose aligned person images}. This has a superior performance to \emph{Enc-Dec rec.~input}, demonstrating the effectiveness of \textbf{alignment}. But they are sub-optimal which may lose information. For example, for an input back-facing person image, such fixed (frontal) pose supervision may mistakenly guide the features to drop the back-facing body information. 3) In contrast, our full aligned texture image as supervision can provide comprehensive and densely semantics aligned information, which results in the best performance.

\textbf{Why not \emph{Directly} use Generated Texture Image for ReID?} How about the performance when the generated  texture images are used as the input for reID? Results show that our scheme significantly outperforms them. The inferior performance is caused by the low quality of the generated texture image (with the texture smoothed/blurred).

\textbf{How does the Quality of Textures affect reID Performance?} We use different backbone networks, \egno, ResNet-101, DenseNet-121, \etc, to train the pseudo texture generators, and then the generated pseudo textures are used to train our SAN-basic network for reID. We find that using deeper and more complex generators can improve the texture quality, which in turn further boosts the reID performance.

\begin{figure}
  \centerline{\includegraphics[width=1.0\linewidth]{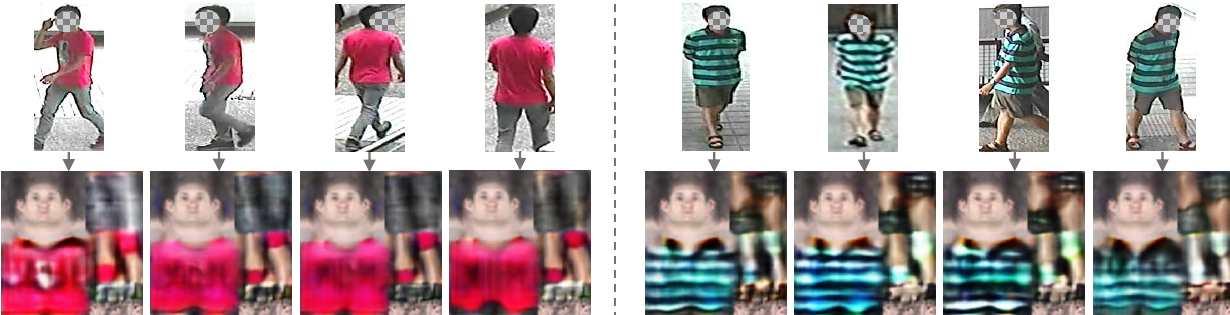}}
  \caption{Two sets of examples of the pairs. Each pair corresponds to the original input image and the generated texture image.} 
  \centering
\label{fig:vis}
\end{figure}

\begin{table*}
  \newcommand{\tabincell}[2]{\begin{tabular}{@{}#1@{}}#2\end{tabular}}
  \centering
  \scriptsize
  \caption{Performance (\%) comparisons with the state-of-the-art methods. Bold numbers denote the best performance, while the numbers with underlines denote the second best.}
  \resizebox{\textwidth}{!}{
    \begin{tabular}{clcccccccccc}
    \toprule
    \multicolumn{2}{c}{\multirow{3}[3]{*}{Method}} & \multicolumn{4}{c}{CUHK03}   & \multicolumn{2}{c}{\multirow{2}[1]{*}{Market1501}} & \multicolumn{2}{c}{\multirow{2}[1]{*}{DukeMTMC-reID}} & \multicolumn{2}{c}{\multirow{2}[1]{*}{MSMT17}} \\
    \multicolumn{2}{c}{} & \multicolumn{2}{c}{Labeled} & \multicolumn{2}{c}{Detected} & \multicolumn{2}{c}{} & \multicolumn{2}{c}{} & \multicolumn{2}{c}{} \\
\cmidrule{3-12}    \multicolumn{2}{c}{} & Rank-1 & mAP   & Rank-1 & mAP   & Rank-1 & mAP   & Rank-1 & mAP   & Rank-1 & mAP \\
    \midrule
    & IDE (ECCV)~\cite{sun2017beyond}   & 43.8  & 38.9 & - & - & 85.3  & 68.5    & 73.2  & 52.8    &  -     & - \\
    \multirow{8}[2]{*}{\tabincell{c}{Pose\\/Part\\/Mask\\-related}} 
          & MGN (ACMMM)~\cite{wang2018learning} & 68.0  & 67.4  & 66.8  & 66.0 & 95.7  & 86.9  & \textbf{88.7 } & \textbf{78.4 } & -      & - \\
          & AACN (CVPR)~\cite{xu2018attention}  & -      & -      & -      & -  & 85.9  & 66.9  & 76.8  & 59.3   & -      & - \\
          & MGCAM (CVPR)~\cite{song2018mask}     & 50.1  & 50.2  & 46.7  & 46.9  & 83.8  & 74.3  & -      & -    & -      & - \\
          & MaskReID (ArXiv)~\cite{qi2018maskreid}  & -      & -      & -      & -   & 90.0  & 70.3  & 78.9  & 61.9  & -      & - \\
          & SPReID (CVPR)~\cite{kalayeh2018human} & -      & -      & -      & -  & 92.5  & 81.3  & 84.4  & 71.0   & -      & - \\
          & Pose Transfer (CVPR)~\cite{liu2018pose}  & 33.8  & 30.5  & 30.1  & 28.2   & 87.7  & 68.9  & 68.6  & 48.1  & -      & - \\
          & PSE (CVPR)~\cite{sarfraz2017pose} & -      & -      & 30.2  & 27.3   & 87.7  & 69.0  & 79.8  & 62.0  & -      & - \\
          & PN-GAN (ECCV)~\cite{qian2018pose}  & -      & -      & -      & -   & 89.4  & 72.6  & 73.6  & 53.2   & -      & - \\
          & Part-Aligned (ECCV)~\cite{suh2018part} & -      & -      & -      & -   & 91.7  & 79.6  & 84.4  & 69.3   & -      & - \\
          & PCB+RPP (ECCV)~\cite{sun2017beyond}  & 63.7  & 57.5  & -      & -   & 93.8  & 81.6  & 83.3  & 69.2   & -      & - \\
    \midrule
    \multicolumn{1}{c}{\multirow{3}[2]{*}{\tabincell{c}{Attention\\-based}}} 
          & DuATM (CVPR)~\cite{si2018dual}  & -      & -      & -      & -  & 91.4  & 76.6  & 81.8  & 64.6   & -      & - \\
          & Mancs (ECCV)~\cite{wang2018mancs}  & 69.0  & 63.9  & 65.5  & 60.5  & 93.1  & 82.3  & 84.9  & 71.8 & -      & - \\
          & FD-GAN (NIPS)~\cite{ge2018fd}     &-  &-  &- &-   & 90.5 & 77.7  & 80.0  & 64.5 &- &- \\
          & HPM (AAAI)~\cite{fu2019horizontal}  & 63.9   & 57.5 & -  & -    & 94.2   & 82.7   & 86.6 & 74.3 & - & -  \\          
    \midrule
    \multicolumn{1}{c}{\multirow{1}[1]{*}{\tabincell{c}{Semantics}}} & DSA-reID (CVPR)~\cite{zhang2019DSA}  & \underline{78.9}  & \underline{75.2}  & \underline{78.2}  & \underline{73.1}  & \underline{95.7}  & \underline{87.6}  & 86.2  & 74.3  & -      & - \\
    \midrule
    \multicolumn{1}{c}{\multirow{3}[2]{*}{\tabincell{c}{Others}}} & GoogLeNet (CVPR)~\cite{wei2018person}  & -  & -  & -  & -  & -  & -  & -  & -  & 47.6      & 23.0 \\
          & PDC (CVPR)~\cite{wei2018person}  & -      & -      & -      & -  & -  & -  & -  & -   & 58.0      & 29.7 \\
          & GLAD (CVPR)~\cite{wei2018person}  & -  & -  & -  & -  & -  & -  & -  & - & 61.4      & 34.0 \\
    \midrule    
    \multirow{1}[2]{*}{Ours} & Baseline (ResNet-50)  & 73.7  & 69.8  & 69.7  & 66.1  & 94.1  & 83.2  & 85.9  & 71.8  & \underline{73.8}  & \underline{47.2} \\
          & SAN       & \textbf{80.1} & \textbf{76.4} & \textbf{79.4} & \textbf{74.6}   & \textbf{96.1} & \textbf{88.0} & \underline{87.9}      & \underline{75.5}      & \textbf{79.2} & \textbf{55.7} \\
    \bottomrule
    \end{tabular}%
    } 
  \label{tab:sto}%
\end{table*}%

\subsection{Comparison with State-of-the-Arts}
Table \ref{tab:sto} shows the performance comparisons of our proposed SAN with the state-of-the-art methods. Our scheme SAN achieves the best performance on CUHK03, Market1501, and MSMT17. It consistently outperforms the approach \emph{DSA-reID} \cite{zhang2019DSA} which also considers the dense alignment. On the DukeMTMC-reID dataset, \emph{MGN} \cite{wang2018learning} achieves better performance, however, it ensembles the local features of multiple granularities and the global features.

\subsection{Visualization of Generated Texture Image}
For the different images with varied poses, viewpoints, or scales, we find the generated texture images from our SAN are well semantically aligned (see Figure \ref{fig:vis}).

\subsection{Partial Person ReID}

Partial person reID is more challenging as the misalignment problem is more severe, where two partial person images are generally not spatially semantics aligned and usually have less overlapped semantics. We also demonstrate the effectiveness of our scheme on the challenging partial person reID datasets of Partial REID \cite{zheng2015partial} and Partial-iLIDS \cite{he2018deep}.

\begin{figure}[h]
  \centerline{\includegraphics[width=1.0\linewidth]{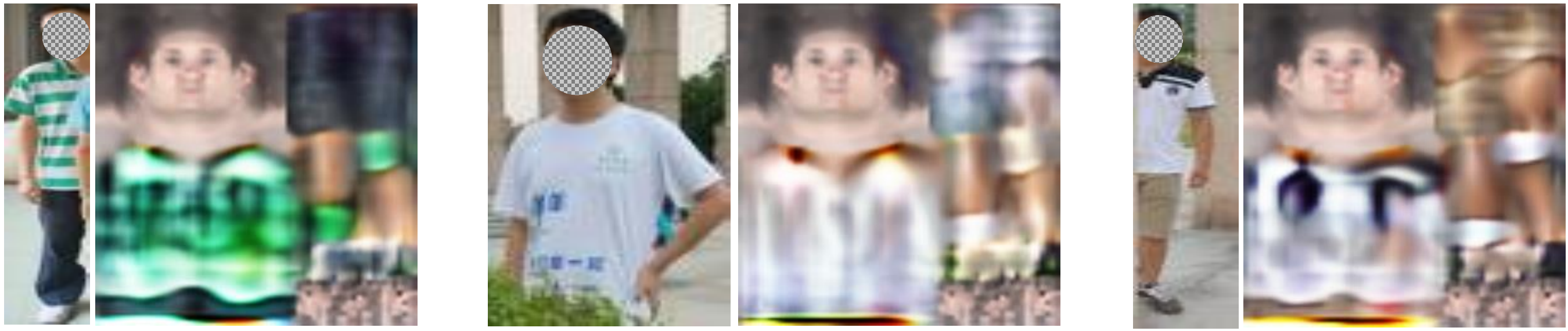}}
  \caption{Three example pairs of (input image, regressed texture images by our SAN) from the Partial REID dataset.} 
  \centering
\label{fig:partial}
\end{figure}

Benefiting from the \emph{aligned full} texture generation capability, our SAN exhibits outstanding performance. Figure \ref{fig:partial} shows our regressed texture images from the SA-Dec are semantically aligned across images even though the input images have severe misalignment.

\begin{table}[h]
  \centering
  \scriptsize
  \tabcolsep=3pt
  \caption{Partial person reID performance on the datasets of Partial REID and Partial-iLIDS (partial images are used as the probe set and holistic images are used as the gallery set). ``*'' means that the network is fine-tuned with holistic and partial person images from Market1501.}
    \begin{tabular}{ccccccc}
    \toprule
    \multirow{2}[4]{*}{Model} & \multicolumn{3}{c}{Partial REID} & \multicolumn{3}{c}{Partial-iLIDS} \\
\cmidrule{2-7}          & Rank-1    & Rank-5    & Rank-10    & Rank-1    & Rank-5    & Rank-10 \\
    \midrule
    AMC+SWM & 36.0    & -      & -      & 49.6  &  -     & - \\
    DSR (single-scale)* & 39.3  &  -     &  -     & 51.1  &  -     & - \\
    DSR (multi-scale)*  & 43.0    &  -     &  -     & \textbf{54.6} & -      & - \\
    \midrule
    Baseline (ResNet-50) & 37.8  & 65.0    & 74.5  & 42.0    & 65.5  & 73.2 \\
    SAN   & 39.7  & 67.5  & 80.5  & 46.9  & 71.2  & 78.2 \\
    \midrule
    Baseline (ResNet-50)*  & 38.9	& 67.7	& 78.2   & 46.1	  & 69.6	 & 76.1
 \\
    SAN*  & \textbf{44.7} & \textbf{72.4} & \textbf{86.0} & 53.7  & \textbf{77.4} & \textbf{81.9} \\
    \bottomrule
    \end{tabular}%
  \label{tab:partial}%
\end{table}%

Table \ref{tab:partial} shows the experimental results. Note that we train SAN on the Market1501 dataset \cite{zheng2015scalable} and test on the partial datasets. We directly take the trained model for Market1501 for testing, \ieno, Baseline (ResNet-50), SAN. In this case, the network seldom sees partial person data. Similar to \cite{he2018deep}, we also fine-tune with the holistic and partial person images cropped from Market1501 (marked by *). SAN* outperforms \emph{Baseline*}, AMC+SWM \cite{zheng2015partial} and is comparable with the state-of-the-art partial reID method DSR \cite{he2018deep}. SAN* outperforms Baseline (ResNet-50)* by 5.8\%, 4.7\%, 7.8\% on Rank-1, Rank-5, and Rank-10 respectively on the Partial REID dataset, and by 7.6\%, 7.8\%, 5.8\% on Rank-1, Rank-5, and Rank-10 respectively on the other Partial-iLIDS dataset. Even without fine-tune, our SAN also significantly outperforms the baseline.

\section{Conclusion}

In this paper, we proposed a simple yet powerful Semantics Aligning Network (SAN) for learning semantics-aligned feature representations for efficient person reID, under the joint supervisions of person reID and semantics aligned texture generation. At the decoder, we add \triIDC over the feature maps as the perceptual loss to regularize the learning. We have synthesized a Paired-Image-Texture dataset (PIT) to train a SAN-PG model, with the purpose to generate pseudo groundtruth texture images for the reID datasets, and to train the SAN.  Our SAN achieves the state-of-the-art performances on the datasets CUHK03, Market1501, MSMT17, and the Partial REID, without increasing computational cost in inference.

\section{Acknowledgments}
This work was supported in part by NSFC under Grant 61571413, 61632001.

{
\small
\bibliographystyle{aaai}
\bibliography{reference}
}

\end{document}